\documentclass[journal]{IEEEtranTIE}

\usepackage{times}
\usepackage{amsmath}
\usepackage{amssymb}
\usepackage{booktabs}
\usepackage{multirow}
\usepackage{graphicx}
\usepackage{float}
\usepackage{stfloats}
\usepackage{url}
\usepackage{textcomp}
\usepackage[table,dvipsnames]{xcolor}
\usepackage[caption=false,font=normalsize,labelfont=sf,textfont=sf]{subfig}
\usepackage[font={scriptsize}]{caption}
\usepackage[ruled,vlined,linesnumbered]{algorithm2e}
\usepackage[utf8]{inputenc}
\usepackage{balance}
\usepackage{cite}
\usepackage{pgfplots}
\usepackage{contour}
\usepackage[hidelinks]{hyperref}
\pgfplotsset{compat=1.18}
\contournumber{24}
\contourlength{.04em}
\SetAlFnt{\footnotesize}
\SetAlgoCaptionSeparator{.}

\renewcommand{\baselinestretch}{0.982}

\title{RRTrack: Robust and Recoverable Object 6D Pose Tracking for Dynamic Scenes}
\author{Junyue Li\textsuperscript{1,2}, Ye Zheng\textsuperscript{2}, Yifan Chen\textsuperscript{2,3}, Zhe Sun\textsuperscript{2}, and Xuelong Li\textsuperscript{2},~\IEEEmembership{Fellow, IEEE}%
\thanks{\textsuperscript{1}College of Computer Science and Technology, Zhejiang University, Hangzhou, China (e-mail: kevin\_li\_win@zju.edu.cn).}
\thanks{\textsuperscript{2}Institute of Artificial Intelligence (TeleAI), China Telecom, China (e-mail: zhengye@westlake.edu.cn; sunzhe@nwpu.edu.cn; xuelong\_li@nwpu.edu.cn).}
\thanks{\textsuperscript{3}College of Future Information Technology, Fudan University, Shanghai 200433, China (e-mail: chenyifan@fudan.edu.cn).}
\thanks{Zhe Sun and Xuelong Li are co-corresponding authors.}
}
\begin{document}

\maketitle
\begin{abstract}
Robust object 6D pose tracking is critical for robotic systems operating in dynamic and occluded scenes. Per-frame estimators are accurate but computationally expensive, while current trackers struggle with fast motion and complete occlusion due to their reliance on continuous visibility. To address these challenges, we present RRTrack, an efficient, recoverable object 6D pose tracker that enables robust tracking through fast motion and target disappearance--reappearance. RRTrack introduces a 2D--6D closed-loop tracking strategy that integrates memory-based video object segmentation (VOS) with 6D pose refinement. The 2D branch maintains target localization, and the 6D branch verifies geometric consistency before memory updates. In addition, a DINOv2-based dual-bank template matching module is developed to recover lost targets by jointly exploiting offline synthetic templates and online observation anchors while maintaining real-time efficiency. We also introduce a synthetic RGB-D benchmark comprising three robotic scenarios with fast motion and full occlusion. Experimental results on the synthetic benchmark demonstrate that RRTrack improves equal-subset mean ADD-S AR by 66.3\% and ADD-S AUC by 65.7\% over FoundationPose while achieving 55.2 FPS. Real-world experiments further validate the robustness of RRTrack under noisy sensing conditions. Project page: \url{https://github.com/7kevin24/RRTrack}.
\end{abstract}


\IEEEpeerreviewmaketitle

\vspace{-16pt}

\section{Introduction}

\IEEEPARstart{A}{ccurate} and real-time object 6D pose tracking is a fundamental perception capability for intelligent robotic systems, enabling applications such as manipulation~\cite{zhang2022practical}, visual servoing~\cite{comport2006real}, and mobile navigation~\cite{kong2025socially}. Compared with single-frame pose estimation, 6D pose tracking continuously estimates object poses over time and provides stable spatial information for downstream planning and control. In practical industrial environments, however, robots frequently encounter rapid object motion, abrupt viewpoint changes, and temporary target disappearance caused by manipulator motion, workspace constraints, or dynamic platforms. These factors often lead to tracking failure and task interruption. Therefore, developing a robust and recoverable 6D pose tracking framework remains an open challenge for industrial robotic systems.

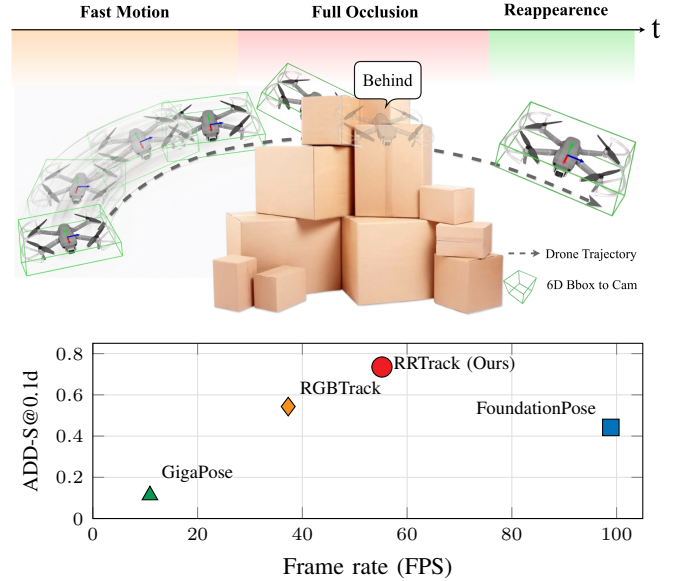
\begin{figure}[!t]
    \centering
    \includegraphics[width=\columnwidth]{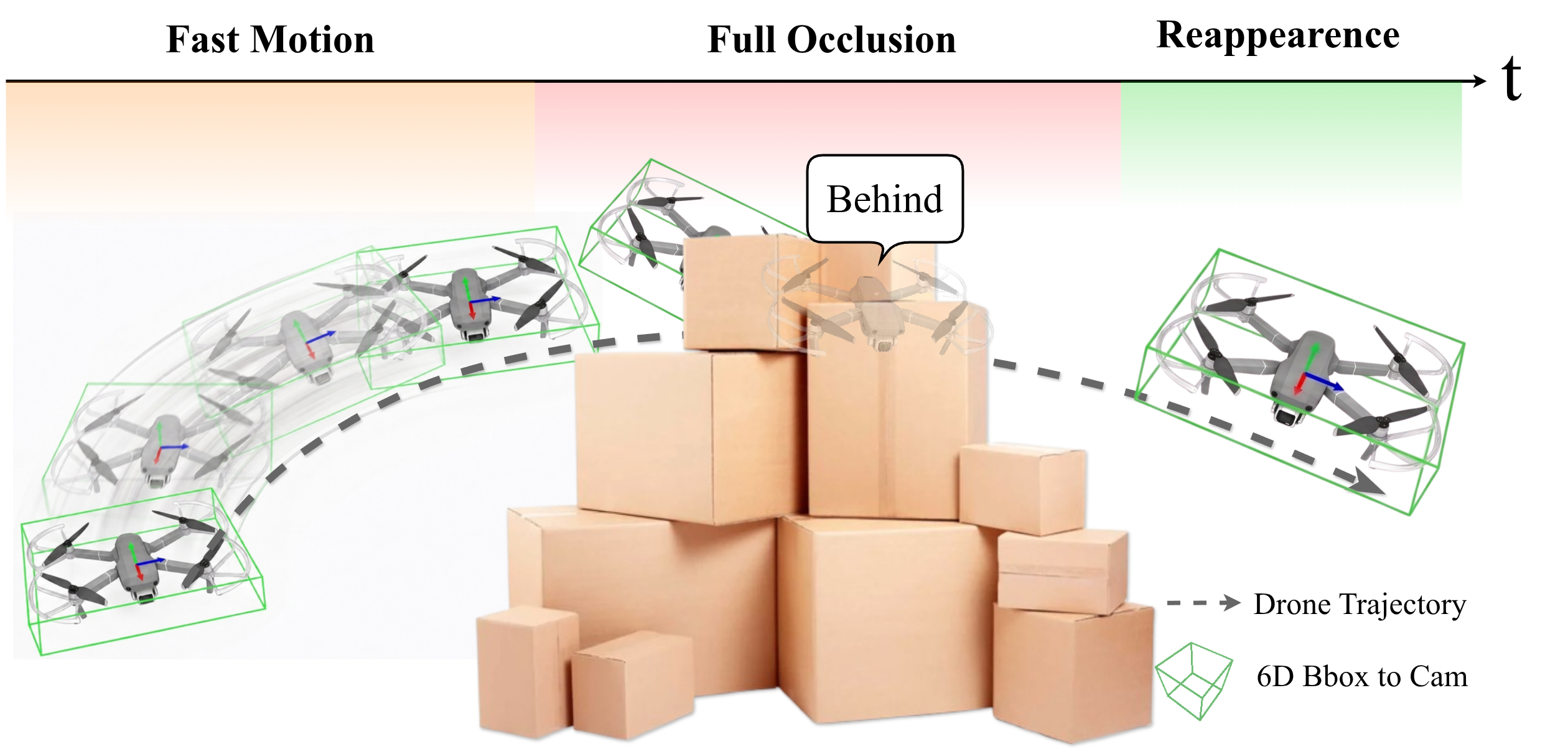}
    \vspace{-1mm}

    \hspace*{-3mm}\begin{tikzpicture}
        \begin{axis}[
            width=\columnwidth,
            height=3.9cm,
            xlabel={\small Frame rate (FPS)},
            ylabel={\fontsize{8pt}{9.6pt}\selectfont ADD-S@0.1d},
            xmin=0,
            xmax=105,
            ymin=0,
            ymax=0.85,
            xtick={0,20,40,60,80,100},
            ytick={0,0.2,0.4,0.6,0.8},
            ymajorgrids=true,
            xmajorgrids=true,
            grid style={line width=.1pt, draw=gray!25},
            axis lines=box,
            only marks,
            tick label style={font=\scriptsize},
            label style={font=\scriptsize},
            clip=false,
        ]
        \addplot[mark=square*, mark size=3.1pt, mark options={fill=RoyalBlue, draw=black}] coordinates{(98.9,0.442)} node[above left, xshift=-2pt, yshift=1pt] {\scriptsize \contour{white}{FoundationPose}};
        \addplot[mark=triangle*, mark size=3.4pt, mark options={fill=ForestGreen, draw=black}] coordinates{(10.9,0.113)} node[above right, xshift=1pt, yshift=1pt] {\scriptsize \contour{white}{GigaPose}};
        \addplot[mark=diamond*, mark size=3.5pt, mark options={fill=BurntOrange, draw=black}] coordinates{(37.3,0.543)} node[above right, xshift=1pt, yshift=1pt] {\scriptsize \contour{white}{RGBTrack}};
        \addplot[mark=*, mark size=3.8pt, mark options={fill=Red, draw=black}] coordinates{(55.2,0.735)} node[right, xshift=1pt, yshift=1pt] {\scriptsize \contour{white}{RRTrack (Ours)}};
        \end{axis}
    \end{tikzpicture}
    \vspace{-5pt}
    \caption{Motivation and performance overview of the proposed RRTrack. Top: the target undergoes fast camera-relative motion, becomes fully occluded, and then reappears with a large displacement, requiring online loss detection and pose recovery. Bottom: equal-subset mean comparison on the proposed synthetic benchmark, where RRTrack improves tracking accuracy while achieving 55.2 FPS.}
    \vspace{-8pt}
    \label{fig:intro_motivation}
\end{figure}

Recent advances in CAD-model-based pose estimation have significantly improved the generalization capability toward unseen objects~\cite{liu2026survey}. Methods such as FoundationPose~\cite{wen2024foundationpose} and GigaPose~\cite{nguyen2024gigapose} leverage render-and-compare refinement to achieve strong pose estimation performance. Building upon these developments, existing tracking approaches have gradually evolved from pose-only tracking to hybrid 2D--6D tracking frameworks. Recently, methods such as RGBTrack~\cite{guo2025rgbtrack} and DynamicPose~\cite{liang2025dynamicpose} introduced image-space tracking cues to enhance robustness under rapid motion.

Despite these advances, current methods still suffer from two major limitations. First, pose-based tracking heavily depends on local optimization and may fail when the object moves outside the convergence basin of the pose refiner. Second, although 2D tracking can maintain object support under fast motion, it often lacks geometric verification and may gradually drift during occlusion or appearance changes. Once unreliable observations are incorporated into memory updates or pose refinement, errors accumulate and eventually cause irreversible tracking failure. Moreover, most existing approaches focus on tracking continuity while providing limited capability for recovering from disappearance--reappearance events. Consequently, robust tracking and reliable recovery remain largely disconnected processes.

Addressing this problem involves three unique challenges. First, rapid inter-frame motion may invalidate pose-conditioned local refinement. Second, prolonged occlusion can contaminate visual memory and induce long-term drift. Third, disappearance--reappearance events require efficient pose re-initialization without sacrificing real-time performance. These challenges highlight the need for a unified framework that jointly considers tracking robustness, memory reliability, and recovery efficiency.

To address these challenges, we propose RRTrack, a training-free and recoverable object 6D pose tracking framework for unseen objects. RRTrack establishes a closed-loop interaction between memory-based video object segmentation and CAD-model-based pose refinement. Specifically, a memory-based segmentation module maintains object support under rapid motion, while a frozen FoundationPose refiner continuously updates object poses from RGB-D observations. To suppress drift accumulation, we further introduce a rendered-mask agreement mechanism that evaluates the consistency between propagated segmentation masks and rendered pose masks. This mechanism regulates memory updates, verifies tracking reliability, and triggers corrective pose updates when inconsistencies occur.

Furthermore, RRTrack incorporates a recovery-oriented tracking strategy to handle disappearance--reappearance events. A DINOv2-based dual-bank recovery module jointly exploits offline synthetic templates and online observation anchors to generate candidates, which are subsequently refined and verified through geometric consistency checks. An adaptive threshold controller further stabilizes transitions among tracking, correction, lost, and recovery states. By tightly coupling visual memory, geometric verification, and retrieval-based recovery, RRTrack achieves robust long-term tracking without manual re-initialization.

Existing 6D pose benchmarks mainly emphasize per-frame accuracy or relatively mild tracking conditions~\cite{hodan2018bop}, leaving limited coverage of the dynamic and heavily occluded scenes targeted in this work. Therefore, we introduce a synthetic benchmark consisting of 3 robotics scenarios, covering 38 video--object tracking episodes and 72 lost--reappear events across 22 RGB-D videos. We further conduct real-world agile-drone experiments for qualitative validation. On the synthetic benchmark, RRTrack improves the equal-subset mean ADD-S AR by 66.3\% and ADD-S AUC by 65.7\% compared with FoundationPose, while running at 55.2 FPS.

Our contributions are summarized as follows:
\begin{itemize}
    \item A 2D--6D closed-loop object 6D pose tracker that unifies temporal visual memory and geometric pose estimation, using rendered-mask agreement to validate observations and protect memory updates.
    \item A DINOv2-based dual-bank recovery strategy with state-aware management for efficient pose re-initialization after target loss.
    \item A synthetic benchmark for evaluating 6D pose tracking under fast motion, full occlusion, and reappearence in robotic settings.
\end{itemize}

\section{Related Work}
\label{sec:related}

\noindent\textbf{Model-based pose estimation}
Classical 6D pose estimators are instance-level or category-level methods~\cite{liu2026survey}, which require object-specific training data, category priors, or retraining for each deployment scenario~\cite{yu2025categorypose}. Such assumptions weaken their generalization and out-of-the-box deployment ability in robotic settings. Recent model-based methods relax this requirement by assuming only the object mesh at test time and building pose hypotheses from rendered or matched object evidence. Template-based methods retrieve similar rendered views or learned descriptors to obtain coarse poses~\cite{okorn2021zephyr, cai2022ove6d, nguyen2024gigapose}. Render-and-compare methods refine candidate poses by comparing rendered RGB-D observations with the input, as in MegaPose and FoundationPose~\cite{labbe2022megapose, wen2024foundationpose}. Correspondence- and flow-based methods instead recover pose from 2D--3D, RGB-D, partial-to-partial, keypoint, or optical-flow matches, improving robustness to clutter and partial observation~\cite{moon2024genflow, lin2024sam6d, zheng2024keypoint}. These methods have improved accuracy and generalization ability for novel-object estimation.

\noindent\textbf{Training-free pose estimation.}
Training-free methods provide an alternative to synthetic data-driven pose estimators by using frozen foundation models for feature extraction, localization, or geometric matching without task-specific finetuning. Self-supervised vision-transformer descriptors, especially DINO/DINOv2 features, have been used for template retrieval~\cite{ausserlechner2024zs6d, ornek2024foundpose}, object localization~\cite{nguyen2023cnos, sam}, correspondence matching~\cite{caraffa2024freeze, poiesi2023gedi}, and coarse pose search~\cite{wang2024object, von2024diffusion}. SAM and its efficient variants provide promptable zero-shot segmentation priors that help localize novel objects before pose estimation~\cite{sam, nguyen2023cnos}, while DINOv2 is often used for transferable semantic and geometric features without object-specific training~\cite{oquab2023dinov2, ornek2024foundpose, caraffa2024freeze}. FoundPose and ZS6D match observations with synthetic templates using frozen visual representations, while FreeZe combines visual and geometric models for zero-shot pose estimation~\cite{ornek2024foundpose, ausserlechner2024zs6d, caraffa2024freeze, aiflow2026,aiflowedge2025,chen2026gentrans}. These works reflect the broader use of vision and vision-language foundation models as reusable priors for optical image processing and downstream visual tasks~\cite{xiao2025optical}. However, they mainly use retrieval or correspondence as a pose-estimation component for the current query, whereas RRTrack embeds such training-free features into a state-aware recovery process for long-horizon tracking.

\noindent\textbf{6D pose tracking.}
Object 6D pose tracking exploits temporal cues for efficient, smooth, and consistent pose estimates. Classical methods update the previous pose through iterative alignment or geometric correspondence, while multi-frame systems such as BundleTrack accumulate observations to jointly optimize object geometry or pose~\cite{li2018deepim, deng2021poserbpf, stoiber2022iterative, wen2021bundletrack}. Recent coarse-to-fine pipelines initialize from the previous pose and apply a local refiner. This is efficient when inter-frame motion remains within the refinement basin but fails once the local prior becomes unreliable~\cite{wen2024foundationpose}. To improve dynamic tracking, RGBTrack combines a 2D tracker, Kalman filtering, and a state machine, while DynamicPose uses depth-informed 2D tracking with motion compensation and filtering~\cite{guo2025rgbtrack, liang2025dynamicpose, chen2026gentrans}. Although image-space tracking preserves target support when pose-only propagation fails, masks or boxes may drift, shrink, or follow distractors during occlusion. RRTrack therefore treats VOS as a 2D observation stream verified by rendered 6D geometry before memory writeback. This closed-loop design enables geometry-gated memory updates, and retrieval-based recovery for long-horizon efficient tracking.

\section{Method}

\begin{figure*}[t]
    \centering
    \includegraphics[width=0.94\textwidth]{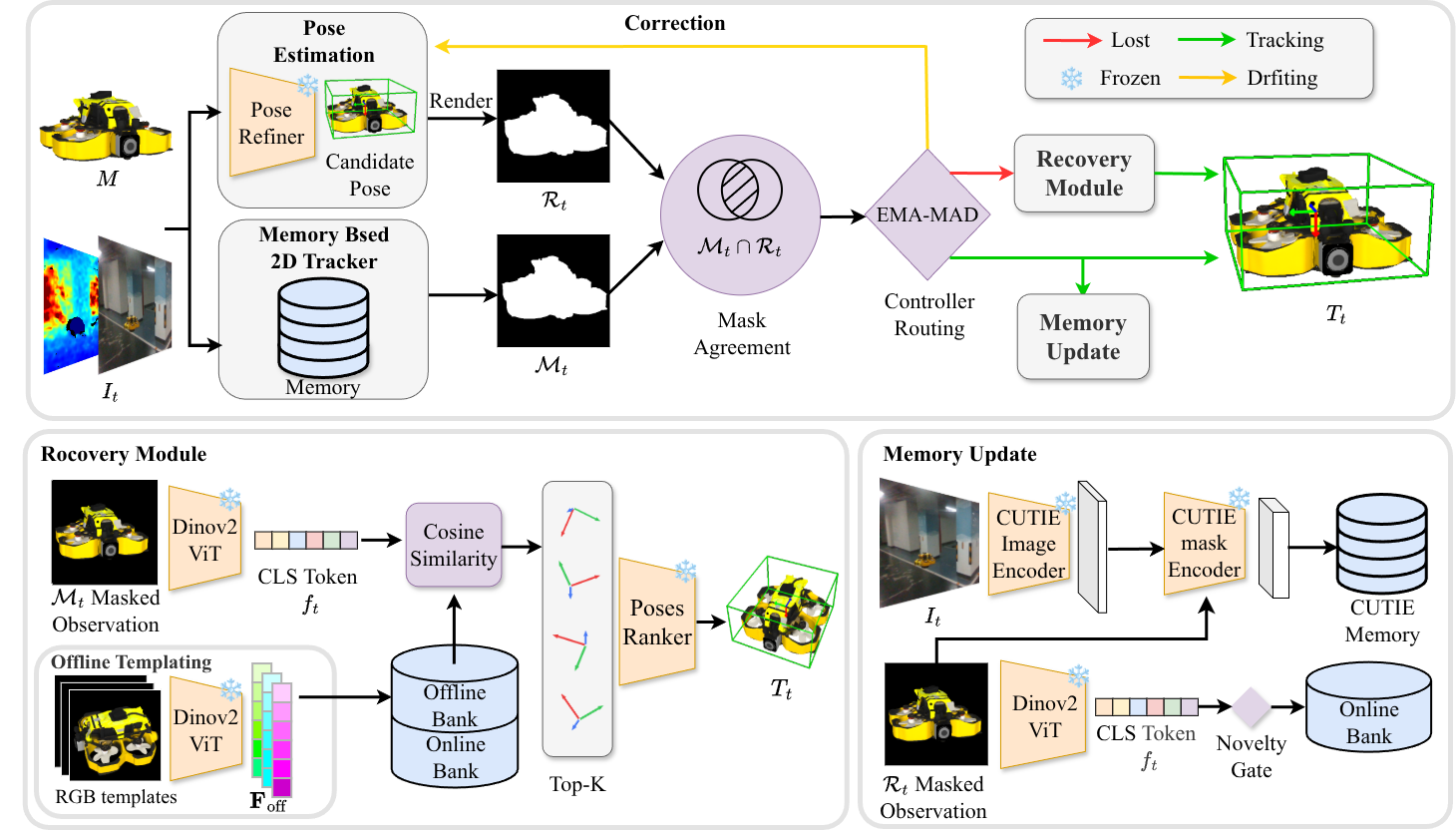}
    \caption{Overview of RRTrack. Given an RGB--D stream, VOS propagates the object mask $\mathcal{M}_t$ and the refiner updates the 6D pose $T_{t}$. The pose-rendered mask $\mathcal{R}_t$ is then compared with $\mathcal{M}_t$. Their agreement, together with the VOS confidence $H_t$, controls state transitions and quality-gated updates to VOS memory and the online template bank. When local pose drift occurs but the mask remains reliable, snap correction re-estimates translation from the mask center and depth. When tracking failure is indicated by low mask agreement, insufficient mask support, or pose stagnation, RRTrack switches to recovery. The recovery module retrieves Top-$K$ coarse pose hypotheses separately from an offline templates bank and an online observation anchors bank. After viewpoint compensation is applied to the offline hypotheses, all candidates are refined and ranked. Finally, the highest-ranked candidate is verified by mask agreement before being accepted.}
    \vspace{-8pt}
    \label{fig:pipeline_overview}
\end{figure*}

RRTrack consists of two main components, as shown in Fig.~\ref{fig:pipeline_overview}. The first is a closed-loop 2D--6D tracking, which couples a memory-based 2D tracker with a frozen FoundationPose refiner. The second is a DINOv2-based dual-bank recovery, which maintains offline synthetic templates and online observation anchors after tracking loss.

\subsection{Closed-Loop 2D--6D Tracking}

The closed-loop tracker maintains object support in 2D, verifies the propagated support through 6D rendered-mask agreement, and uses the verified geometry to control correction and memory updates. This design allows the 2D and 6D streams to complement each other: the 2D stream provides a stable observation cue, while the 6D stream filters unreliable masks before they are commited.

\paragraph{Rendered-mask agreement for tracking assessment}
The closed loop requires a memory-based 2D observation module that can maintain object support across short occlusions and appearance changes. We use CUTIE~\cite{cheng2024cutie} as this module to propagate an object-centric mask over time. Given the current RGB frame and the maintained memory, the module outputs a propagated mask $\mathcal{M}_t$ and a confidence signal derived from the prediction entropy. Given the current pose estimate $T_t$, we render a binary object mask $\mathcal{R}_t$ and compare it with $\mathcal{M}_t$ using mask agreement scores
\begin{equation}
    P_t = \frac{|\mathcal{M}_t \cap \mathcal{R}_t|}{|\mathcal{M}_t|}, \qquad
    S_t = \frac{|\mathcal{M}_t \cap \mathcal{R}_t|}{|\mathcal{R}_t|},
\end{equation}
where $P_t$ measures how well the rendered pose explains the propagated 2D support, and $S_t$ measures how much of the rendered object remains covered by the propagated mask. We further compute the average foreground entropy $H_t$ of the VOS prediction as an entropy-based confidence signal. $P_t$, $S_t$, and $H_t$ characterize whether the current frame is both geometrically consistent and visually reliable.

\paragraph{Snap correction for local drift}
In our dynamic setting, fast inter-frame motion can make local 6D refinement fail even when the propagated mask remains usable. We therefore trigger a local snap update when precision falls below the tracking threshold. When the propagated-mask area remains sufficiently large, RRTrack uses the 2D observation to re-estimate translation before full recovery is required. We denote this trigger by
\begin{equation}
    I_{\mathrm{snap}} = \mathbb{I}\big(P_t < \tau_{P}^{\mathrm{track}} \;\wedge\; |\mathcal{M}_t| \ge a_{\mathrm{snap}}\big),
\end{equation}
where $\tau_{P}^{\mathrm{track}}$ is the tracking-continuation threshold supplied by an exponential moving average--median absolute deviation (EMA--MAD) adaptive threshold manager. The manager updates EMA and MAD statistics only on stable tracking frames and derives metric-specific thresholds from these robust running statistics. The term $a_{\mathrm{snap}}$ denotes the minimum propagated-mask area required for local correction. We adopt the current 2D observation to construct a translation-only proposal. Let $\mathbf{c}(\mathcal{M}_t)$ denote the 2D box center of $\mathcal{M}_t$, and let $z_t$ denote the median valid depth from $D_t$ inside the mask region. We back-project this visual center into the camera frame as
\begin{equation}
    \hat{\mathbf{t}}_t = \Pi^{-1}(\mathbf{c}(\mathcal{M}_t), z_t; K),
\end{equation}
where $\Pi^{-1}(\cdot;K)$ denotes back-projection from a 2D point and depth to camera-frame 3D coordinates using the camera intrinsic matrix $K$. The proposal keeps the previous rotation unchanged and only overwrites translation,
\begin{equation}
    \hat{T}_t = [R_{t-1} \mid \hat{\mathbf{t}}_t].
\end{equation}
This proposal is then passed to the local 6D refiner. The snap operation helps correct moderate drift by realigning the pose with the current 2D support before complete tracking failure.

\paragraph{Quality-gated memory injection}
The two agreement scores separate different writeback conditions. A high $P_t$ indicates that the propagated mask is largely explained by the current rendered pose, whereas a high $S_t$ indicates that the rendered object is sufficiently covered by the propagated observation. When $P_t$ is high but $S_t$ is low, the observation may correspond to an occluded or under-segmented object region. Storing it as a persistent anchor would bias later memory retrieval toward partial views. We therefore use stricter $P_t$--$S_t$ gates for long-term memory and online-bank updates, and a separate confidence gate for short-term VOS memory.

The same agreement signals are used to decide whether the current observation should be written back into persistent memory, and the online template bank. We use $\tau_*$ for precision, entropy, and other decision thresholds, and reserve $\rho_*$ for support thresholds based on $S_t$. All thresholds are supplied by the EMA--MAD manager from stable-tracking statistics. We express the main writeback gates as
\begin{equation}
    I_{\mathrm{mem}}^{\mathrm{long}} = \mathbb{I}\big(P_t > \tau_{P}^{\mathrm{long}} \;\wedge\; S_t > \rho_{\mathrm{long}}\big),
\end{equation}
\begin{equation}
    I_{\mathrm{mem}}^{\mathrm{short}} = \mathbb{I}\big(P_t > \tau_{P}^{\mathrm{short}} \;\wedge\; H_t < \tau_H\big),
\end{equation}
where $I_{\mathrm{mem}}^{\mathrm{long}}$ writes the rendered pose mask $\mathcal{R}_t$ into long-term memory, while $I_{\mathrm{mem}}^{\mathrm{short}}$ allows the propagated mask $\mathcal{M}_t$ to update short-term memory. The online template bank is updated only under a separate geometric-quality gate,
\begin{equation}
    I_{\mathrm{bank}} = \mathbb{I}\big(P_t > \tau_{P}^{\mathrm{bank}} \;\wedge\; S_t > \rho_{\mathrm{bank}}\big),
\end{equation}
so that newly added anchors are both visually plausible and pose-consistent. This geometry-filtered writeback serves as a closed-loop gate: pose verifies the mask before the mask is stored, which suppresses drift accumulation in the 2D observation module.

\subsection{State-Aware Recovery}

The recovery module is activated only when the tracker enters an unreliable state. It first uses explicit loss and stagnation tests to decide when local refinement should stop, then retrieves coarse pose hypotheses from complementary offline and online template banks, and finally accepts a recovered pose only after geometric validation against the current 2D observation.

\paragraph{State logic}
The controller switches between local refinement and retrieval-based recovery according to explicit loss and stagnation tests. During normal tracking, local pose refinement remains active only if the rendered-mask precision stays above an adaptive floor and the propagated-mask area remains sufficiently large. We declare the target lost via
\begin{equation}
    I_{\mathrm{lost}} = \mathbb{I}\big(
    P_t < \tau_{P}^{\mathrm{floor}}
    \;\vee\;
    |\mathcal{M}_t| < a_{\mathrm{lost}}
    \big),
\end{equation}
where $\tau_{P}^{\mathrm{floor}}$ is the precision-floor threshold obtained from the same EMA--MAD manager used in closed-loop tracking, and $a_{\mathrm{lost}}$ denotes the minimum propagated-mask area required to keep local tracking active.

To prevent the tracker from remaining trapped in poor local minima without complete loss, we explicitly monitor short-term pose stagnation. Given the estimated pose $T_t=[R_t\mid \mathbf{t}_t]$ and a temporal window $W_{\mathrm{stag}}$, we compute the relative rotation angle $\Delta \theta_t = \angle(R_t, R_{t-W_{\mathrm{stag}}})$ and translation magnitude $\Delta d_t = \|\mathbf{t}_t - \mathbf{t}_{t-W_{\mathrm{stag}}}\|_2$. Stagnation is identified via
\begin{equation}
    I_{\mathrm{stag}} = \mathbb{I}\big(
    \max\!\left(\frac{\Delta \theta_t}{\theta_{\mathrm{th}}}, \frac{\Delta d_t}{d_{\mathrm{th}}}\right) < 1
    \;\wedge\;
    P_t < \tau_{P}^{\mathrm{stag}}
    \big),
\end{equation}
where $\theta_{\mathrm{th}}$ and $d_{\mathrm{th}}$ are fixed kinematic thresholds, and $\tau_{P}^{\mathrm{stag}}$ is the precision-stagnation threshold obtained from the same EMA--MAD manager. This criterion captures frames in which the pose changes little over time despite persistently low agreement, a characteristic signature of local optimizer stagnation. When $I_{\mathrm{lost}} \vee I_{\mathrm{stag}} = 1$, local refinement is disabled and retrieval-based recovery is invoked.

\paragraph{Dual-bank feature retrieval}
To support robust re-initialization under severe occlusion, we propose a dual-bank formulation consisting of two distinct pose-feature components. An extensive offline bank $\mathcal{B}_{\mathrm{off}}$ maintains rendered templates to provide broad viewpoint coverage without introducing onboard computational overhead. Concurrently, a lightweight online bank $\mathcal{B}_{\mathrm{on}}$ dynamically incorporates real observation anchors during high-quality tracking to bridge the domain gap.

\begin{equation}
    \mathcal{B}_{m} = \left\{(\mathbf{f}_i^{m}, T_i^{m})\right\}_{i=1}^{N_m}, \quad m \in \{\mathrm{off}, \mathrm{on}\}.
\end{equation}

During recovery, the masked observation of the current frame is passed through DINOv2~\cite{oquab2023dinov2} to extract an $L_2$-normalized global CLS token, $\mathbf{f}_t = \Phi_{\mathrm{CLS}}(I_t \odot \mathcal{M}_t)$. We independently retrieve the top-$K_{\mathrm{ret}}$ candidates from the offline and online banks. Each bank is evaluated with its own EMA--MAD validation threshold, preventing dominance of online scores while preserving fallback hypotheses from the synthetic bank. For each bank $m \in \{\mathrm{off},\mathrm{on}\}$, we define the candidate index set as
\begin{equation}
    \mathcal{C}_t^m = \mathrm{TopK}(\{\cos(\mathbf{f}_t,\mathbf{f}_i^m)\}_{i=1}^{N_m}, K_{\mathrm{ret}}), \qquad m \in \{\mathrm{off}, \mathrm{on}\},
\end{equation}
where $\cos(\mathbf{f}_t,\mathbf{f}_i^m)$ denotes the cosine similarity between $L_2$-normalized descriptors, and $\mathrm{TopK}(\{\cos(\mathbf{f}_t,\mathbf{f}_i^m)\}, K_{\mathrm{ret}})$ returns the indices of the $K_{\mathrm{ret}}$ highest-scoring templates in bank $m$. The global CLS-token descriptor narrows the coarse orientation hypotheses without the computational overhead of explicit dense correspondence search.

\paragraph{Candidate solve and acceptance}
Offline templates are rendered in a canonical centered configuration. Their retrieved poses, together with observation-derived online anchors, are passed to FoundationPose's standard RoIGrid-based refinement, which re-establishes the current-frame pose from the RGB--D observation and mask. Each retrieved pose $T_i$ then defines a candidate initialization
\begin{equation}
    \tilde{T}_{t,i}=T_i, \qquad i \in \mathcal{C}_t^{\mathrm{off}} \cup \mathcal{C}_t^{\mathrm{on}},
\end{equation}
where $T_i$ is the retrieved offline or online pose. These candidates are refined once and scored by the frozen FoundationPose refiner and ranker~\cite{wen2024foundationpose}. Let $T_t^\star$ denote the refined hypothesis with the highest FoundationPose ranker score, and define its rendered-mask precision as $P_t^\star$. We accept recovery via
\begin{equation}
    I_{\mathrm{acc}} = \mathbb{I}\big(P_t^\star > \tau_{P}^{\mathrm{track}}\big),
\end{equation}
where $P_t^\star = \frac{|\mathcal{M}_t \cap \mathcal{R}(T_t^\star)|}{|\mathcal{M}_t|}$ denotes the rendered-mask precision of the refined hypothesis against the current observation, and $\tau_{P}^{\mathrm{track}}$ is obtained from the same EMA--MAD manager used in closed-loop tracking. Otherwise, the tracker remains in the lost state. If retrieval provides no acceptable candidate, the implementation falls back to a spherical pose search.

\section{Experiments}
\subsection{Experiment Settings}

We evaluate RRTrack through two complementary setups. First, we construct a synthetic RGB-D benchmark in Isaac Sim to quantify pose accuracy, recovery behavior, and processing speed using simulator-generated ground-truth 6D object poses. Second, we provide a visual qualitative comparison on real-world agile-drone sequences to assess temporal consistency and 6D bounding-box alignment beyond simulation, since precise 6D ground truth is unavailable for the physical flights.

\paragraph{Implementation details}
We use $640\times320$ RGB-D streams, downsample CUTIE to 1/4 resolution, and extract DINOv2 ViT-S/14 CLS-token descriptors at $224\times224$ with L2 normalization and 15\% context padding. During recovery, we independently retrieve top-$K{=}3$ candidates from the offline and online banks and apply distinct EMA--MAD validation thresholds to the two banks. The offline template bank contains 256 rendered full-sphere views with 180$^\circ$ in-plane augmentation, while the online bank stores up to 64 observation anchors with FIFO replacement. The adaptive threshold manager uses EMA coefficient $\alpha=0.05$ and a 100-frame MAD buffer. Benchmark and code will be released publicly. All experiments are conducted on a single RTX 4090 GPU under a unified evaluation setting.

\paragraph{Metrics}
For all synthetic subsets, we report ADD and ADD-S with the same normalized distance threshold $0.1d$: area under accuracy-threshold curve (AUC) is integrated up to $0.1d$, and average recall (AR) is computed at the $0.1d$ criterion~\cite{Xiang2018-dv,hodan2018bop}.
For recovery-focused analysis, we additionally report success rate and post-recovery ADD/ADD-S AUC and AR. For each annotated lost--reappear event, we examine the ten frames after the target reappears. A recovery is counted as successful if at least one frame in this interval reaches ADD-S AR greater than 0.9. For the successfully recovered events, post-recovery ADD/ADD-S AUC and AR are computed over the ten-frame window after the first successful recovery frame. Unrecovered events contribute zero to the event-level post-recovery scores. This event-level formulation separates whether a method can re-establish a valid pose from how well the recovered pose remains geometrically consistent immediately afterward. Averaging only raw frame accuracy after reappearance would conflate early unstable outputs with sustained recovery quality.

\subsection{Evaluation on Synthetic Benchmarks}
We therefore evaluate synthetic RGB-D streams that explicitly include viewpoint changes, fast motion, occlusion, disappearance, and reappearance. Standard BOP benchmarks primarily evaluate 6D pose estimation or tracking under relatively mild dynamics, whereas our synthetic benchmark explicitly covers high dynamics and disappearance--reappearance events in the target dynamic tracking regime.

\paragraph{Dataset}
The synthetic benchmark contains three scenario families generated in Isaac Sim. The simulator provides synchronized RGB-D streams, masks, camera intrinsics, and simulator-generated ground-truth object poses, enabling the same quantitative evaluation protocol across all subsets.

\noindent\textbf{Franka-YCB.}
The Franka-YCB subset evaluates pick-and-place manipulation with a Franka arm and multiple YCB objects, recording RGB-D streams, masks, camera intrinsics, and object poses. Each scene contains two to four pickable objects, and the camera follows an orbiting view with height sampled from 0.85--1.50 m and orbit arc from 45$^\circ$--180$^\circ$. During grasping and placing, the arm frequently passes between the camera and the target object, while the orbiting camera adds viewpoint changes. These factors create partial and full occlusions during repeated pick-and-place interaction.

\noindent\textbf{Agile quadcopter.}
The agile quadcopter subset evaluates high-speed flight through procedural ring-gate courses with linear, orbital, and figure-eight layouts. Each run contains 800--1000 frames, and the controller uses a maximum speed of 7.0 m/s and acceleration of 15.0 m/s$^2$. Agile motion produces large inter-frame displacement and can move the target outside the camera field of view before re-entry. The ring gates also introduce many partial occlusions and occasional full occlusions, together with rapid scale and viewpoint variation.

\begin{figure}[!t]
    \centering
    \setlength{\tabcolsep}{1pt}
    \renewcommand{\arraystretch}{0.9}
    \begin{tabular}{ccc}
        \includegraphics[width=0.30\linewidth]{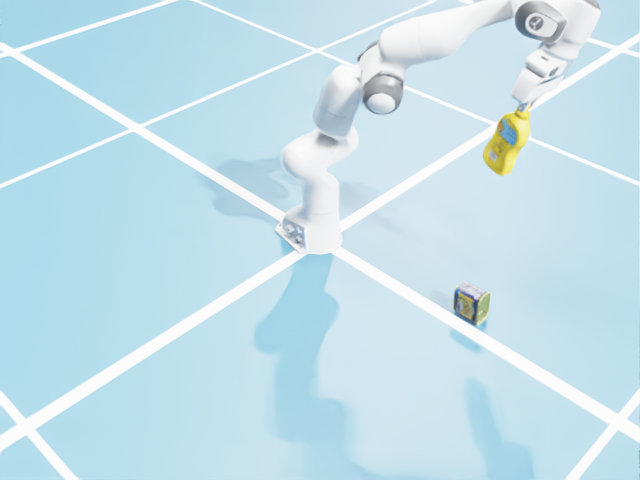} &
        \includegraphics[width=0.30\linewidth]{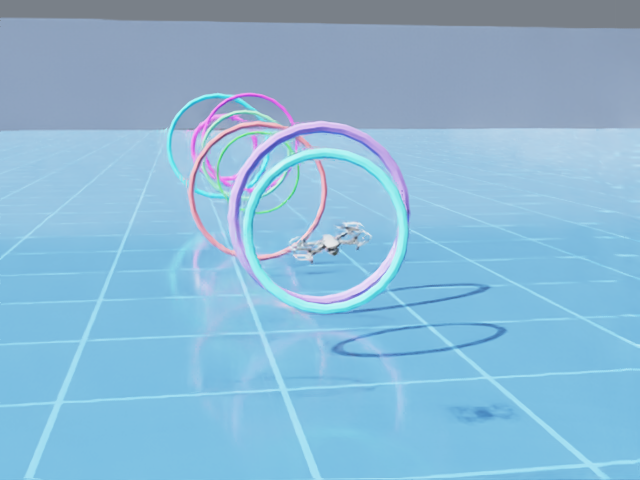} &
        \includegraphics[width=0.30\linewidth]{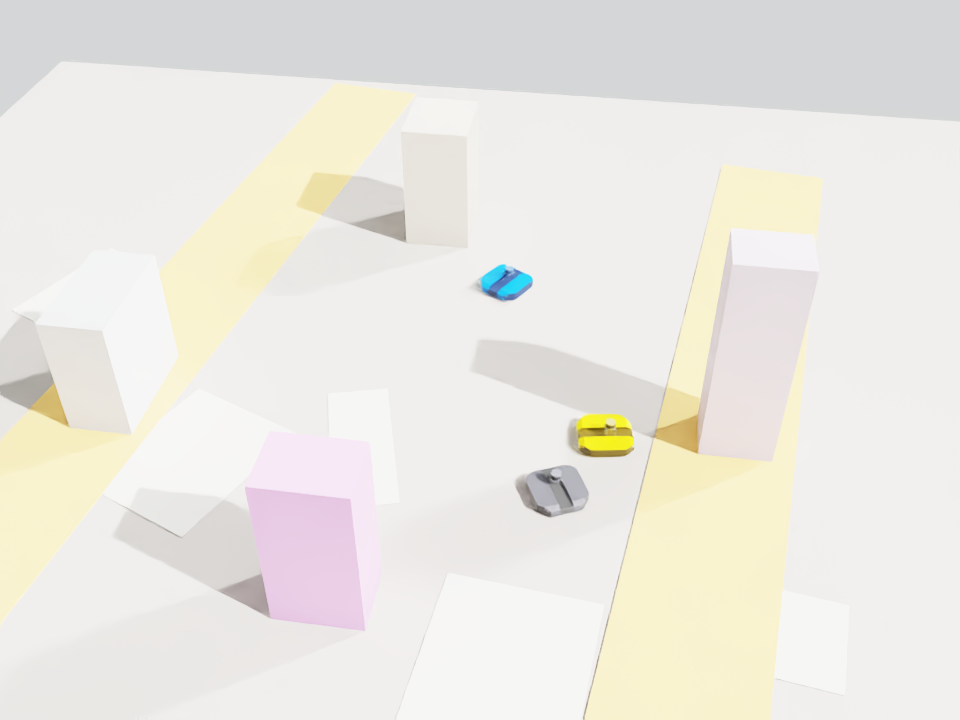} \\
        \includegraphics[width=0.30\linewidth]{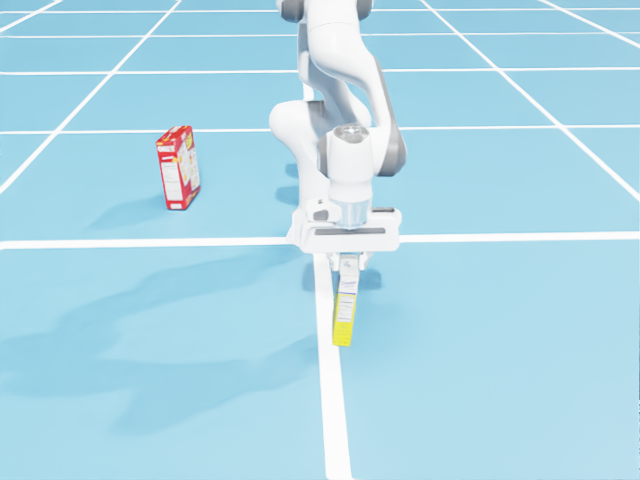} &
        \includegraphics[width=0.30\linewidth]{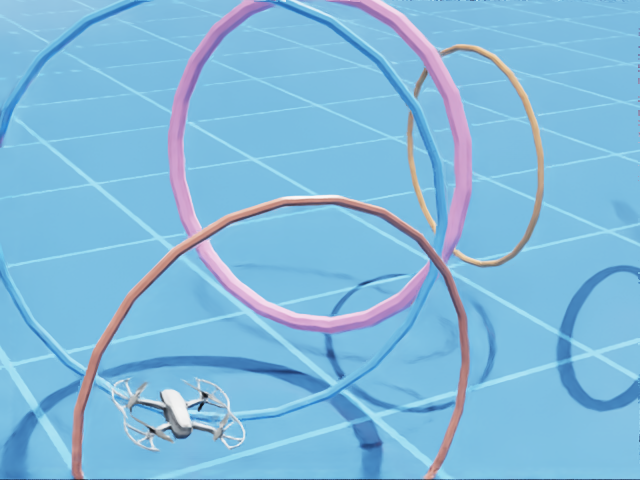} &
        \includegraphics[width=0.30\linewidth]{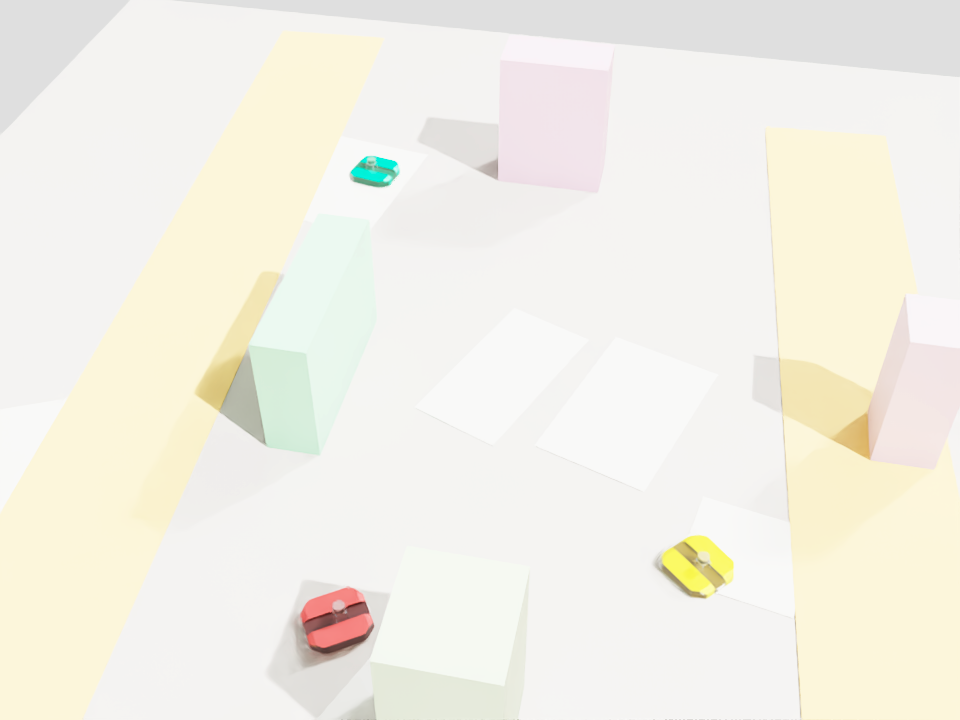} \\
        \includegraphics[width=0.30\linewidth]{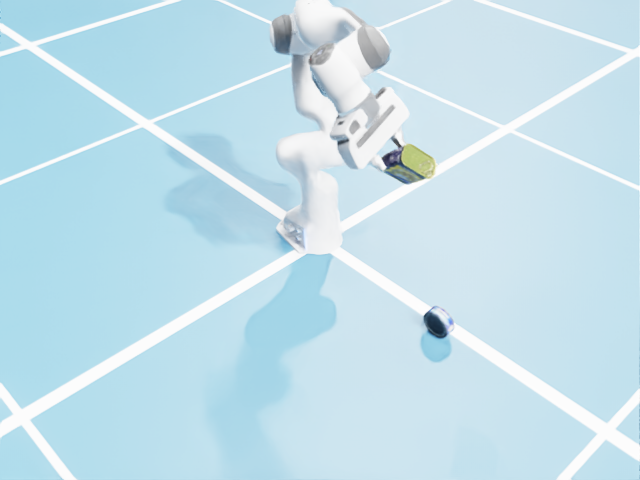} &
        \includegraphics[width=0.30\linewidth]{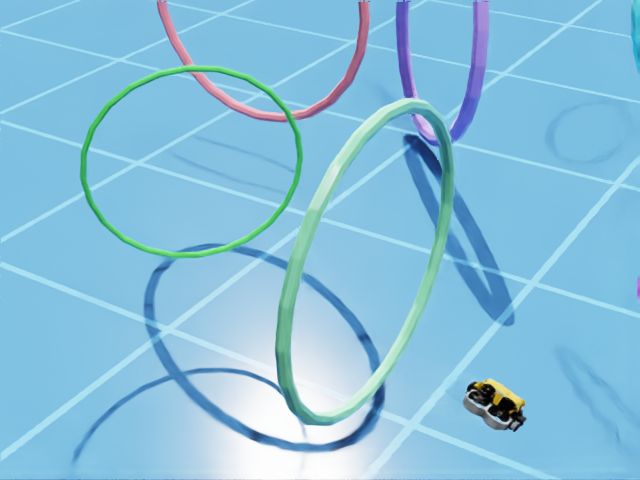} &
        \includegraphics[width=0.30\linewidth]{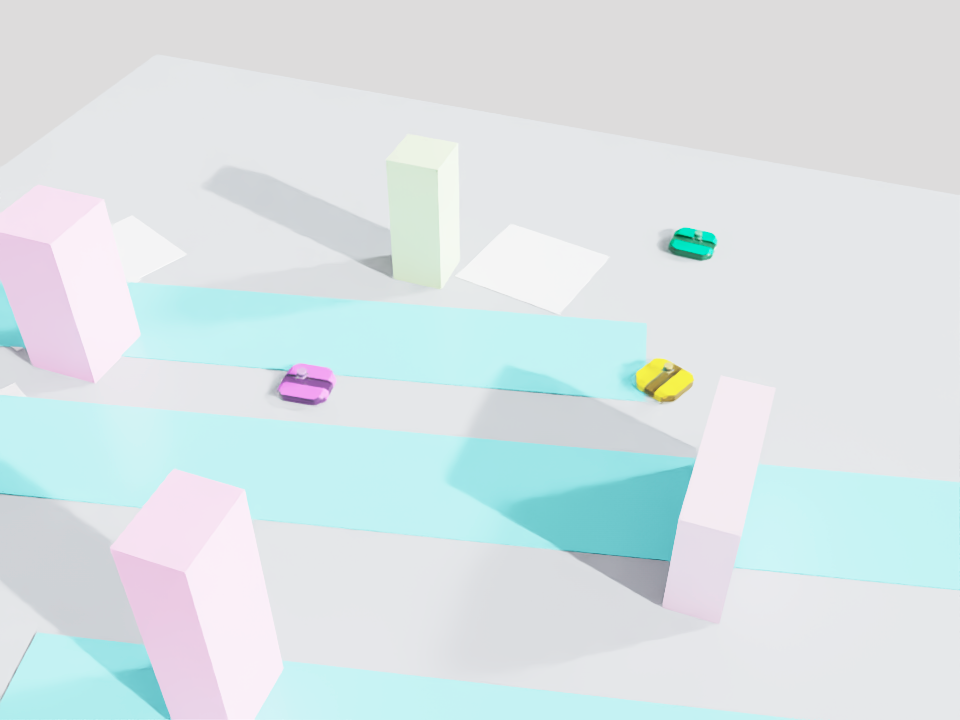} \\
        {\footnotesize Franka} &
        {\footnotesize Agile drone} &
        {\footnotesize Dingo}
    \end{tabular}
    \caption{Demonstration of the proposed synthetic benchmark. Each column presents three sampled frames from one scenario family. The examples illustrate viewpoint change, fast motion, occlusion, and reappearance in the target dynamic tracking regime.}
    \label{fig:dataset_overview}
\end{figure}

\noindent\textbf{Dingo factory.}
The Dingo subset evaluates automated guided vehicle(AGV) navigation in factory-like scenes with multiple colored Dingo robots and randomized obstacles. Each 15 s run contains three robots in a scene with 6.0 m default radius and five to seven rectangular obstacles up to 2.0 meter high. These obstacles repeatedly block the tracked robot, producing full occlusions, extended absence, and delayed recovery cases. Figure~\ref{fig:dataset_overview} presents representative sampled frames from these three scenario families. Table~\ref{tab:synthetic_stats} summarizes the resulting benchmark scale and recovery-event coverage, with 38 video--object tracking episodes and 72 lost--reappear events across 22 videos.

\begin{table}[!htbp]
\centering
\scriptsize
\caption{Synthetic benchmark statistics.}
\label{tab:synthetic_stats}
\renewcommand{\arraystretch}{1.08}
\setlength{\tabcolsep}{3pt}
\resizebox{0.88\columnwidth}{!}{%
\begin{tabular}{lrrrr}
\toprule
Subset & Obj. & Videos & Episodes & Lost--reappear \\
\midrule
Dingo & 6 & 5 & 15 & 28 \\
Franka-YCB & 8 & 5 & 11 & 6 \\
Agile drone & 2 & 12 & 12 & 38 \\
\midrule
Total & 13 & 22 & 38 & 72 \\
\bottomrule
\end{tabular}%
}
\vspace{-12pt}
\end{table}

\paragraph{Compared methods}
We compare RRTrack with methods that can operate in the same model-based, zero-shot RGB-D setting and achieve inference speeds above 10 FPS on continuous video streams. FoundationPose~\cite{wen2024foundationpose} is a unified framework for 6D pose estimation and tracking. In its tracking mode, it efficiently refines the pose from the previous frame using a pose refiner. GigaPose~\cite{nguyen2024gigapose} is originally designed for single-frame pose estimation. For a controlled comparison, we adapt it to a last pose refiner enabling tracking-efficiency. RGBTrack~\cite{guo2025rgbtrack} is included as a recent RGB-D tracking baseline. Together, these methods represent recent state-of-the-art approaches to 6D object tracking.

All methods are evaluated with RGB-D input on the same recorded frame streams. Accuracy is computed over the same evaluated frames for all methods, and FPS is reported only to assess processing efficiency. Table~\ref{tab:main_results} reports the per-object comparison. For each object, ADD and ADD-S are reported as AUC integrated up to $0.1d$ and AR evaluated at the $0.1d$ threshold. Higher is better for all metrics, including FPS. The MEAN row first aggregates each synthetic subset and then averages Franka, agile drone, and Dingo with equal weight. FPS follows the same equal-subset protocol.

\begin{table*}[t]
\centering
\scriptsize
\caption{Comparison with the state-of-the-art methods on 16 objects.}
\label{tab:main_results}
\renewcommand{\arraystretch}{1.08}
\setlength{\tabcolsep}{2.6pt}
\resizebox{\textwidth}{!}{%
\begin{tabular}{cc||cccc|cccc|cccc|cccc}
\toprule
\multirow{3}{*}{Scenario} & \multirow{3}{*}{Object} & \multicolumn{4}{c|}{FoundationPose} & \multicolumn{4}{c|}{GigaPose} & \multicolumn{4}{c|}{RGBTrack} & \multicolumn{4}{c}{RRTrack (Ours)} \\
& & \multicolumn{2}{c}{ADD} & \multicolumn{2}{c|}{ADD-S} & \multicolumn{2}{c}{ADD} & \multicolumn{2}{c|}{ADD-S} & \multicolumn{2}{c}{ADD} & \multicolumn{2}{c|}{ADD-S} & \multicolumn{2}{c}{ADD} & \multicolumn{2}{c}{ADD-S} \\
& & AUC & AR & AUC & AR & AUC & AR & AUC & AR & AUC & AR & AUC & AR & AUC & AR & AUC & AR \\
\midrule
\multirow{8}{*}{Franka} & brick & 24.7 & \textbf{83.3} & \textbf{57.4} & \textbf{98.8} & 0.1 & 0.0 & 0.5 & 1.8 & \textbf{25.5} & \textbf{83.7} & 57.0 & 97.0 & 23.5 & 82.0 & 55.8 & 97.0 \\
 & cracker & \textbf{75.1} & \textbf{97.5} & \textbf{81.7} & \textbf{100.0} & 43.6 & 64.9 & 56.7 & 98.0 & 19.4 & 27.1 & 66.9 & 93.8 & \textbf{75.1} & \textbf{97.5} & \textbf{81.7} & \textbf{100.0} \\
 & sugar & 63.2 & 79.0 & 67.3 & 79.2 & 1.9 & 5.1 & 11.2 & 48.3 & \textbf{63.4} & \textbf{79.1} & \textbf{78.8} & \textbf{96.1} & 62.4 & 78.1 & 67.0 & 79.2 \\
 & soup & 34.9 & 67.2 & 46.9 & 70.0 & 0.7 & 2.4 & 3.3 & 14.4 & 33.5 & 64.7 & 50.6 & 76.9 & \textbf{38.7} & \textbf{75.3} & \textbf{53.2} & \textbf{81.9} \\
 & mustard & 62.2 & 88.7 & 71.6 & 90.0 & 2.5 & 3.7 & 7.0 & 38.6 & 17.8 & 24.2 & 30.6 & 44.7 & \textbf{65.0} & \textbf{93.3} & \textbf{75.2} & \textbf{95.7} \\
 & tuna & 13.0 & 68.6 & 53.4 & 98.3 & 0.8 & 3.4 & 4.9 & 12.3 & \textbf{15.7} & \textbf{75.9} & \textbf{54.8} & \textbf{99.6} & 6.0 & 52.7 & 41.6 & 76.9 \\
 & jell-o & 30.9 & 61.9 & 44.6 & 68.4 & 5.9 & 16.8 & 12.7 & 27.4 & 18.6 & 42.9 & 43.0 & 70.5 & \textbf{47.1} & \textbf{85.6} & \textbf{62.7} & \textbf{94.2} \\
 & spam & \textbf{39.1} & \textbf{61.4} & 46.6 & 63.6 & 0.9 & 2.4 & 2.2 & 5.2 & 30.6 & 44.2 & \textbf{49.8} & \textbf{79.8} & 24.5 & 39.5 & 34.6 & 56.9 \\
\midrule
\multirow{2}{*}{Agile drone} & quad1 & 5.1 & 8.6 & 6.9 & 9.2 & 1.3 & 2.0 & 3.7 & 9.1 & 22.4 & 35.7 & 46.6 & 79.8 & \textbf{23.1} & \textbf{44.3} & \textbf{49.9} & \textbf{81.7} \\
 & quad2 & 5.9 & 8.4 & 6.9 & 9.1 & 1.5 & 3.6 & 6.3 & 15.2 & 19.7 & 32.4 & 33.5 & 65.7 & \textbf{23.6} & \textbf{47.3} & \textbf{48.1} & \textbf{86.3} \\
\midrule
\multirow{6}{*}{Dingo} & black & 15.9 & 33.8 & 27.3 & 40.8 & 0.1 & 0.0 & 0.1 & 0.2 & 0.1 & 0.4 & 2.5 & 7.9 & \textbf{23.9} & \textbf{48.9} & \textbf{42.7} & \textbf{71.4} \\
 & blue & 9.0 & 20.0 & 14.8 & 26.2 & 0.1 & 0.0 & 0.1 & 0.1 & 4.8 & 10.4 & 8.2 & 13.5 & \textbf{10.9} & \textbf{21.5} & \textbf{16.3} & \textbf{26.5} \\
 & purple & 6.3 & 21.3 & 15.8 & 24.8 & 0.1 & 0.7 & 2.1 & 6.5 & 3.1 & 8.8 & 6.4 & 14.0 & \textbf{8.9} & \textbf{22.0} & \textbf{17.3} & \textbf{35.9} \\
 & red & 5.6 & 15.1 & 11.2 & 25.1 & 0.1 & 0.0 & 0.1 & 0.1 & 3.6 & 11.1 & 9.2 & 22.2 & \textbf{10.6} & \textbf{25.6} & \textbf{20.5} & \textbf{37.4} \\
 & teal & 22.9 & 49.3 & 36.6 & 55.6 & 0.1 & 0.0 & 0.1 & 0.1 & 7.2 & 15.7 & 12.9 & 24.9 & \textbf{24.7} & \textbf{51.4} & \textbf{39.9} & \textbf{64.4} \\
 & yellow & 10.2 & 45.3 & 38.6 & 54.3 & 0.1 & 0.0 & 0.1 & 0.0 & 6.7 & 15.3 & 12.4 & 30.5 & \textbf{26.6} & \textbf{55.0} & \textbf{42.0} & \textbf{66.3} \\
\midrule
\multicolumn{2}{c||}{MEAN} & 21.1 & 39.7 & 29.7 & 44.2 & 1.2 & 3.1 & 3.8 & 11.3 & 16.7 & 30.4 & 31.7 & 54.3 & \textbf{31.7} & \textbf{55.1} & \textbf{49.2} & \textbf{73.5} \\
\multicolumn{2}{c||}{FPS (Hz)} & \multicolumn{4}{c|}{\textbf{98.9}} & \multicolumn{4}{c|}{10.9} & \multicolumn{4}{c|}{37.3} & \multicolumn{4}{c}{55.2} \\
\bottomrule
\end{tabular}%
\vspace{-8pt}
}
\end{table*}

\paragraph{Results} RRTrack obtains the highest equal-subset mean accuracy across all four pose metrics, with ADD AUC/AR of 31.7/55.1 and ADD-S AUC/AR of 49.2/73.5. It also runs at 55.2 FPS on the recorded RGB-D streams.

Franka-YCB mainly tests manipulation-induced occlusion. In this relatively smoother regime, several methods remain competitive on individual objects. For example, the strongest entries on brick, sugar, tuna, and spam are shared by different methods. RRTrack is nevertheless strong on soup, mustard, and jell-o while remaining competitive on other objects. This indicates that the proposed coupling is useful beyond complete disappearance cases, improving robustness when partial occlusion and viewpoint change make local pose updates less stable.

The agile-drone subset stresses fast inter-frame displacement, rapid scale and viewpoint change, and out-of-view re-entry. The ADD-S AR gap is large in this high-dynamics setting. On quad1, RRTrack obtains 81.7, compared with 9.2 for FoundationPose, 9.1 for GigaPose, and 79.8 for RGBTrack. On quad2, RRTrack obtains 86.3, compared with 9.1, 15.2, and 65.7. These rows suggest that purely local pose-refinement or single-frame pose-estimation baselines become less stable under agile motion, while RRTrack maintains more robust performance by coupling 2D target support with 6D pose refinement.

The Dingo subset further emphasizes disappearance and delayed recovery. Across all six Dingo objects, RRTrack gives the best ADD and ADD-S scores in Table~\ref{tab:main_results}. The black row gives the clearest example: RRTrack obtains ADD-S AR of 71.4, compared with 40.8 for FoundationPose, 0.2 for GigaPose, and 7.9 for RGBTrack. This result shows that the recovery-oriented design is most beneficial when tracking must continue through object absence and reappearance.

Overall, the table supports the design choices. The largest gains appear in the scenarios where fast motion and lost--reappearance events are central, while the pipeline still preserves real-time throughput on the full benchmark.

The consistent gap between ADD and ADD-S is expected from the evaluated object geometry. Dingo and the agile drone are approximately symmetric at the image resolutions used in our experiments, making some rotations visually ambiguous.

\begin{figure}[!htbp]
    \vspace{-12pt}
    \centering
    \includegraphics[width=\columnwidth]{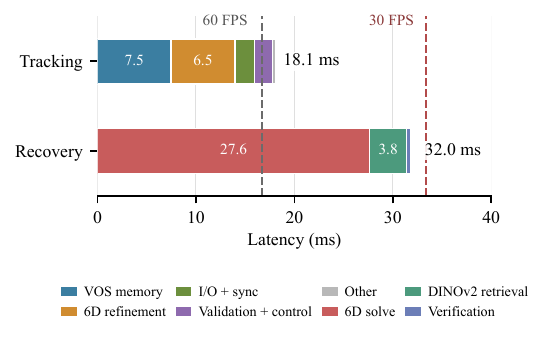}
    \caption{Average runtime of different modules on the synthetic benchmarks. Fine-grained profiler entries are aggregated into semantic module groups, with dashed reference lines marking 60 FPS and 30 FPS latency budgets. Normal tracking is dominated by VOS memory propagation and local 6D refinement, whereas recovery is dominated by batched 6D candidate refinement and ranking.}
    \label{fig:runtime_breakdown}
    \vspace{-6pt}
\end{figure}

\paragraph{Inference speed}
We profile RRTrack on the synthetic benchmarks in Figure~\ref{fig:runtime_breakdown}. The full pipeline averages 18.1 ms per frame (55.2 FPS), remaining well within the 30 FPS processing budget. This confirms that adding validation and recovery does not compromise real-time tracking throughput.

During normal tracking, VOS memory propagation and local 6D refinement take 7.46 ms and 6.50 ms, respectively; all other operations are lightweight, keeping the tracking path close to 60 FPS. Recovery is triggered only after target loss and therefore does not burden normal tracking. A successful recovery frame takes 32.02 ms versus 18.08 ms for normal tracking.  Although the batched candidate search increases latency by 77.1\%, it remains within the 33.33 ms budget for 30 FPS. Thus, RRTrack gains recovery capability without compromising tracking speed.

\subsection{Real-World Drone Experiment}

\begin{figure}[!htbp]
    \vspace{-12pt}
    \centering
    \includegraphics[width=\columnwidth]{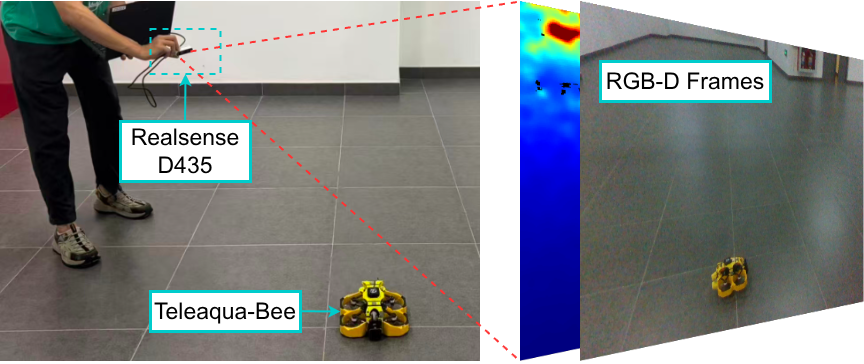}\\[2pt]
    \includegraphics[width=\columnwidth]{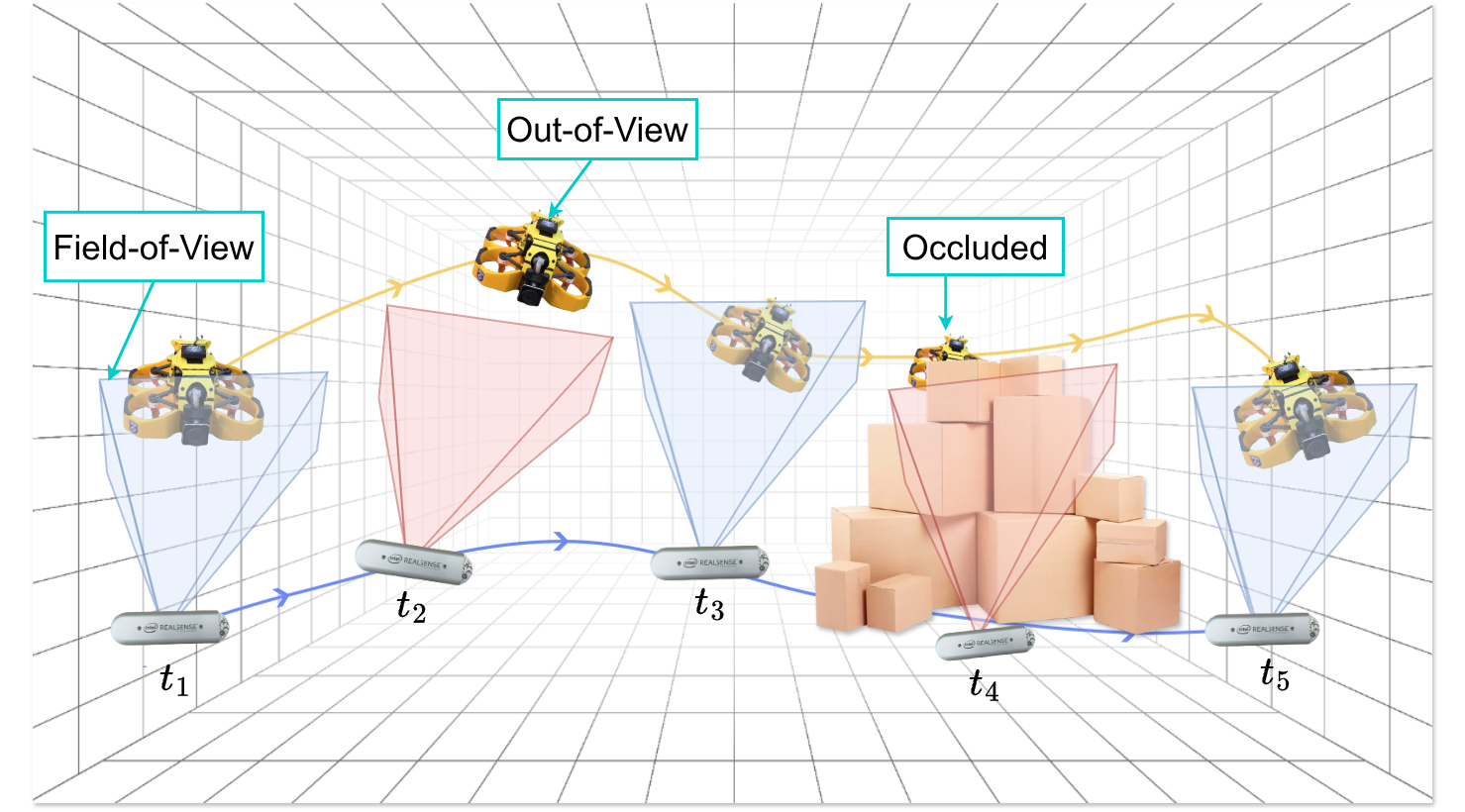}
    \caption{Real-world experimental setups. The upper panel shows RGB-D capture using a handheld Intel RealSense D435 camera at 640$\times$480 pixels and 60~FPS, with the Teleaqua-Bee platform developed by TeleAI as the target. The lower panel shows the target-loss design: handheld-camera tracking latency causes the drone to move out of view, while the drone actively flies behind an obstacle to introduce full occlusion.}
    \label{fig:real_setup}
    \vspace{-16pt}
\end{figure}

We further evaluate RRTrack on recorded real-world agile-drone sequences. Figure~\ref{fig:real_setup} summarizes the real-world experimental setup. RGB-D sequences are captured at a resolution of 640$\times$480 and 60 FPS with a handheld Intel RealSense D435 camera. The camera is operated without a gimbal, introducing continuous shake and intermittent severe jitter. We deliberately include two recovery-critical events: latency in manually following the drone causes it to leave the camera field of view, while the drone actively flies behind an obstacle to introduce full occlusion. These unmodeled sensing conditions and designed target-loss events provide a qualitative stress test of the proposed tracking and recovery design.

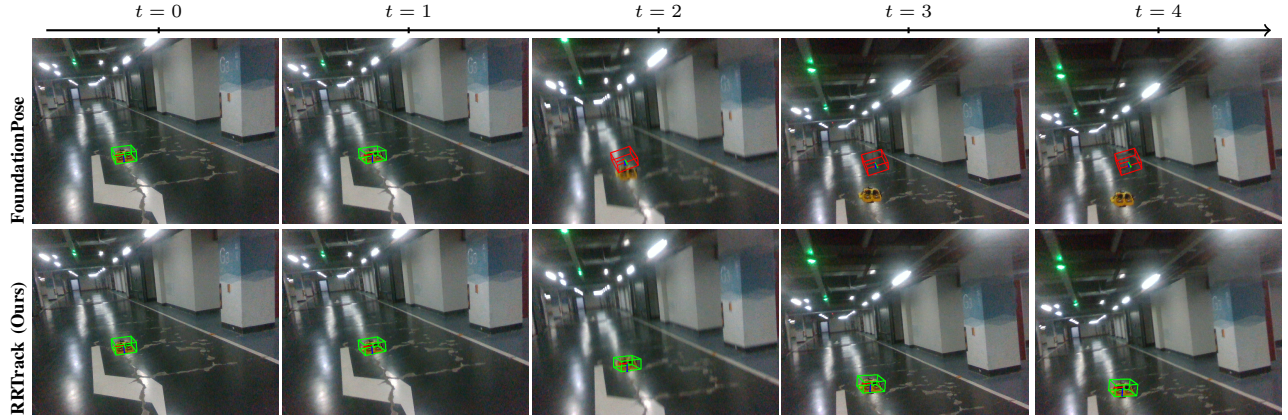
\begin{figure*}[t]
    \vspace{-8pt}
    \centering
    \newlength{\figsixcolw}
    \setlength{\figsixcolw}{0.18\textwidth}
    \setlength{\tabcolsep}{0.6pt}
    \renewcommand{\arraystretch}{0.92}
    \scriptsize
    \begin{tabular}{@{}l@{\hspace{2pt}}ccccc@{}}
        & \multicolumn{5}{c}{%
        \begin{tikzpicture}[x=\dimexpr\figsixcolw+2\tabcolsep\relax,y=1em]
            \path[use as bounding box] (0,-0.25) rectangle (5,0.7);
            \draw[->,thick] (0.05,0) -- (4.95,0);
            \foreach \x/\lab in {0.5/0,1.5/1,2.5/2,3.5/3,4.5/4}{
                \draw[thick] (\x,-0.16) -- (\x,0.16) node[above=1pt] {$t=\lab$};
            }
        \end{tikzpicture}} \\[-0.1em]
        \rotatebox{90}{\textbf{FoundationPose}} &
        \includegraphics[width=\figsixcolw]{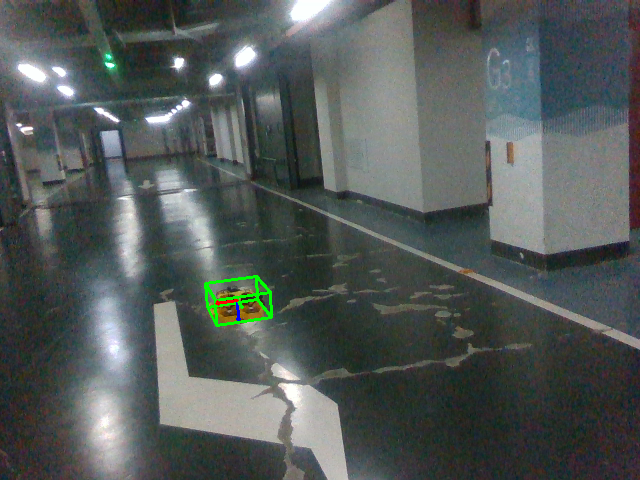} &
        \includegraphics[width=\figsixcolw]{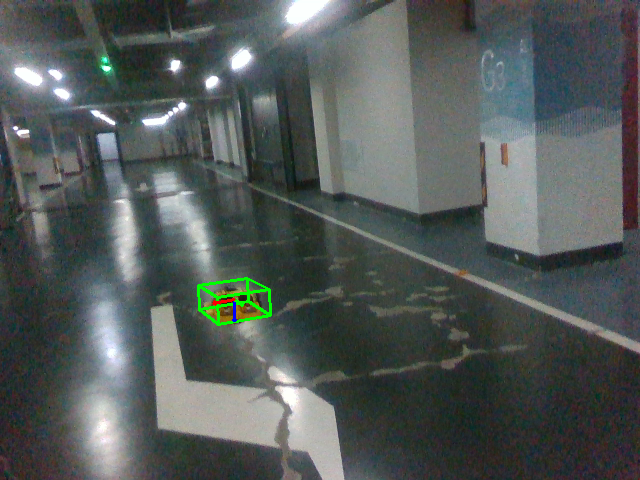} &
        \includegraphics[width=\figsixcolw]{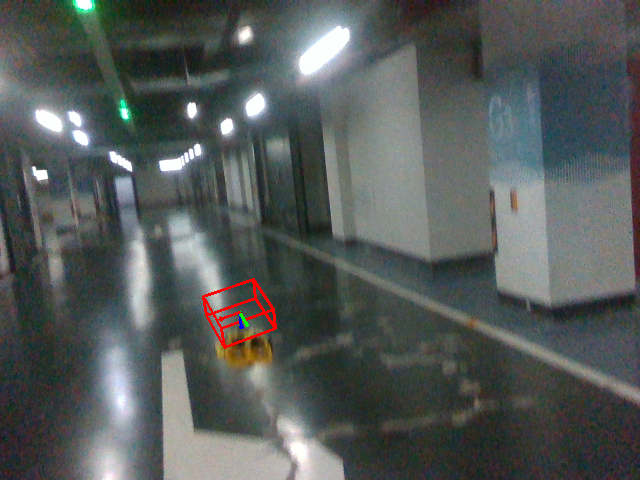} &
        \includegraphics[width=\figsixcolw]{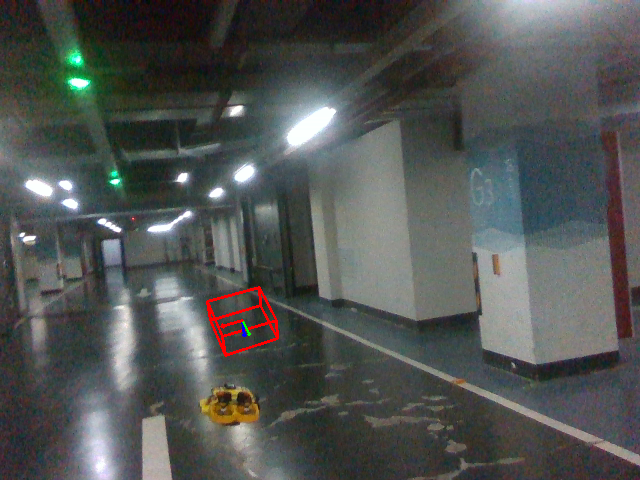} &
        \includegraphics[width=\figsixcolw]{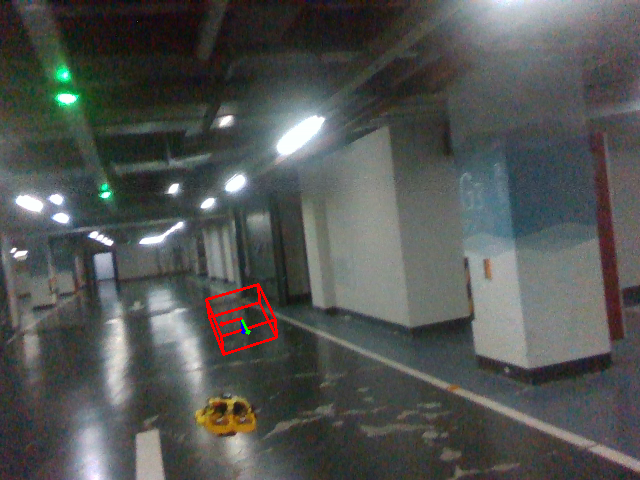} \\
        \rotatebox{90}{\textbf{RRTrack (Ours)}} &
        \includegraphics[width=\figsixcolw]{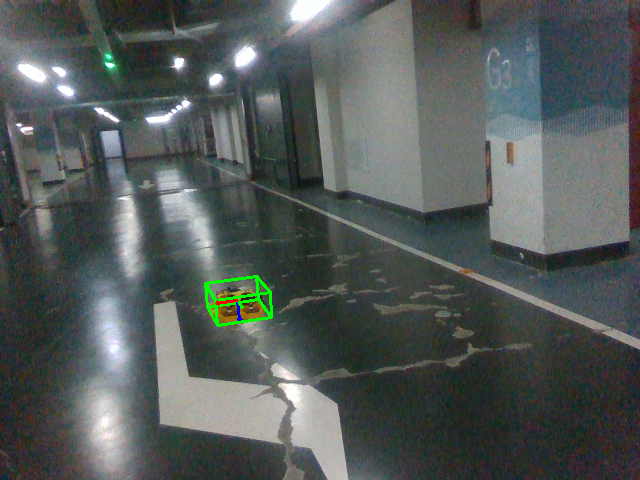} &
        \includegraphics[width=\figsixcolw]{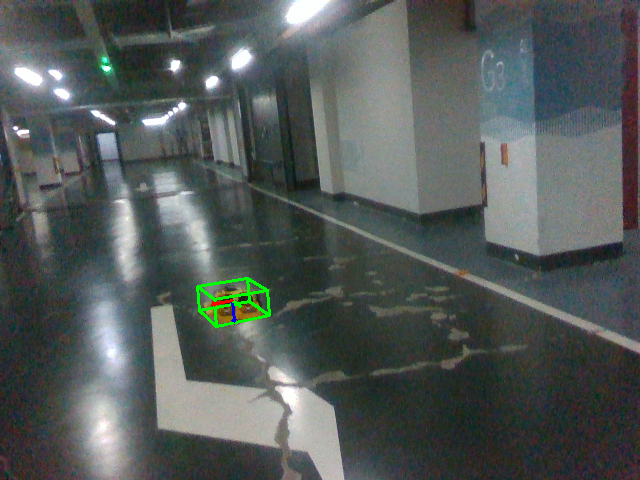} &
        \includegraphics[width=\figsixcolw]{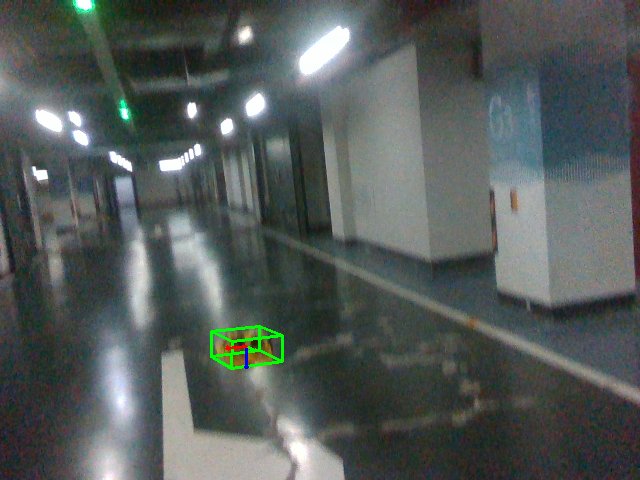} &
        \includegraphics[width=\figsixcolw]{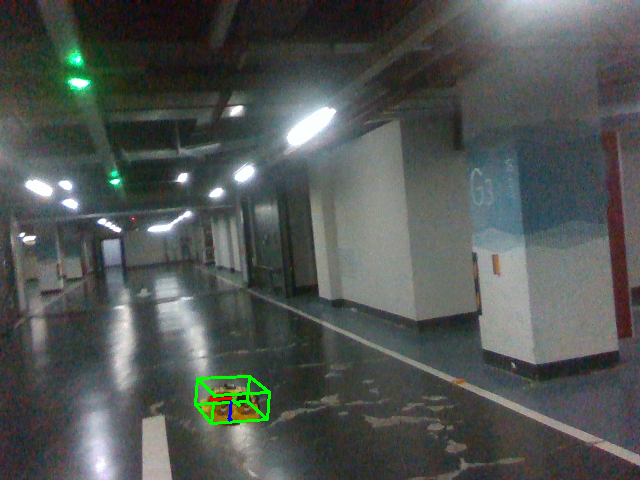} &
        \includegraphics[width=\figsixcolw]{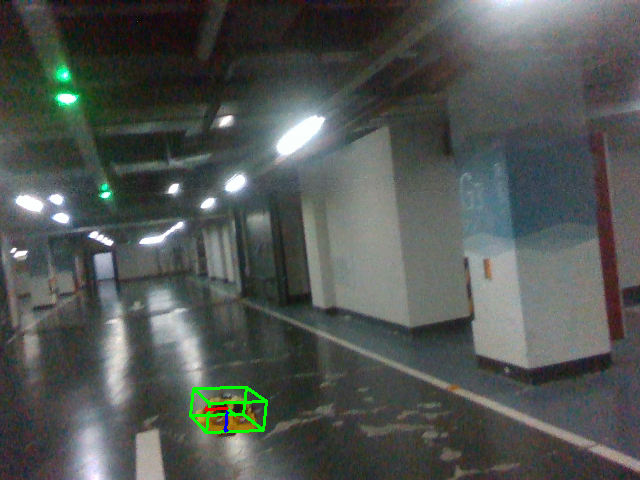}
    \end{tabular}
    \caption{Temporal qualitative comparison between FoundationPose and RRTrack on consecutive frames from the underground parking-garage setting. FoundationPose is initially aligned but drifts under handheld shake, apparent motion, blur, and occlusion, and then loses the target. RRTrack keeps a visually consistent 6D bounding-box overlay across the same frames.}
    \label{fig:real_drone_dynamics}
    \vspace{-12pt}
\end{figure*}

\begin{figure*}[t]
    \centering
    \includegraphics[width=0.95\textwidth]{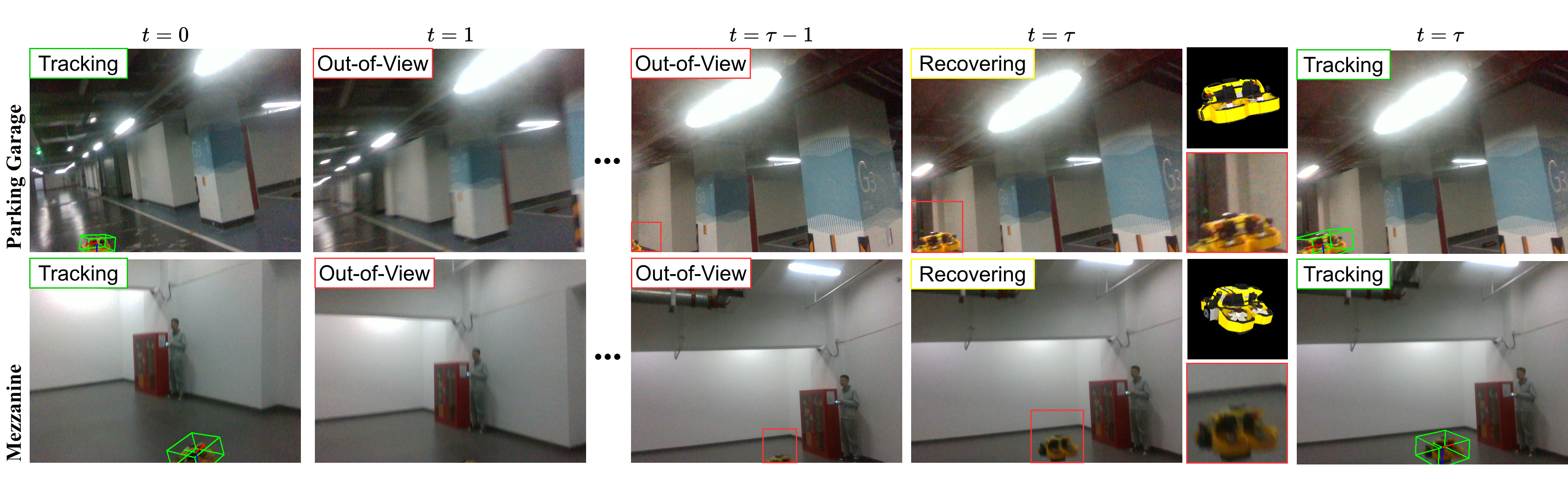}
    \caption{Real-world disappearance--reappearance examples. Each row shows one event: at $t=0$, the target is still visible but about to leave the camera view, from $t=1$ to $t=\tau-1$, the target is fully outside the field of view, and at $t=\tau$, the target reappears. The panels at $t=\tau$ include an enlarged view of the reappearing target and the best-matched template, followed by the recovered 6D bounding-box overlay.}
    \label{fig:real_drone_recovery}
    \vspace{-8pt}
\end{figure*}

The real-world evaluation uses two physical settings. The first is an underground parking garage with dim overall illumination, strong light--dark contrast, noisy video, and frequent occlusions. In this sequence, the drone flies slowly while weaving around pillars, and the handheld D435 follows the target without a gimbal, producing continuous shake and occasional severe jitter. This sequence is used in Figure~\ref{fig:real_drone_dynamics} to examine the 2D--6D tracking design under sustained handheld motion and occlusion. The second setting is an open mezzanine with dim but more uniform illumination. It provides a longer noisy sequence with continuous handheld shake, and intervals in which the target leaves the camera field of view. This sequence is used in Figure~\ref{fig:real_drone_recovery} to examine the recovery module after disappearance and re-entry. Because precise 6D ground truth is not available in these physical scenes, the real-world study is qualitative and focuses on visual temporal consistency and 6D bounding-box alignment in the selected tracking and recovery intervals.

Figure~\ref{fig:real_drone_dynamics} shows continuous tracking on consecutive frames from the underground parking-garage sequence. FoundationPose remains aligned at the beginning of the sequence, but the 6D bounding-box overlay starts to drift as inter-frame viewpoint changes increase. At $t=2$, the drone has a relatively large velocity, and the 60 FPS recording still exhibits noticeable motion blur. FoundationPose therefore immediately loses track. RRTrack instead keeps the target support from the 2D observation stream and maintains a visually consistent 6D poses across the selected frames, suggesting that the memory-based observation design provides a more stable target cue when local refinement becomes unreliable.

Figure~\ref{fig:real_drone_recovery} illustrates disappearance and reappearance in the two real-world settings. The first row shows the underground parking-garage setting, and the second row shows the open-mezzanine setting. At $t=0$, the target is still visible but is about to leave the camera view. From $t=1$ to $t=\tau-1$, the target is fully outside the field of view, so local frame-to-frame pose refinement has no target observation to follow. At $t=\tau$, the target reappears. The associated panels show that RRTrack retrieves the best-matched template from memory and then re-establishes the 6D bounding-box overlay on the full image, reflecting the role of the observation bank in storing target appearances before disappearance and restarting pose estimation after re-entry.

Overall, these real-world experiments provide qualitative evidence that RRTrack remains effective under challenging conditions. The method maintains visually consistent temporal tracking without scene-specific tuning and runs at a practical real-time speed of 49.3 FPS on the recorded real-world sequences.

\subsection{Ablation Study}

We further conduct ablation studies under the synthetic benchmark to examine the contribution of the 2D observation stream and the recovery design.

\subsubsection{Effects of 2D Backbones}

This study replaces the 2D observation backbone to understand the importance of the memory-based VOS stream. Under identical conditions, we replace CUTIE with SAM2~\cite{ravi2024sam} or OSTracker~\cite{ye2022ostrack} while keeping the rest of the protocol unchanged. Table~\ref{tab:ablation_backbone_overall} reports equal-subset ADD/ADD-S AUC and AR over the synthetic benchmark, following the same percentage scale and $0.1d$ threshold convention as Table~\ref{tab:main_results}. FPS follows the same equal-subset protocol, and higher is better for all metrics.

\begin{table}[!htbp]
\centering
\small
\caption{2D observation-backbone ablation.}
\label{tab:ablation_backbone_overall}
\renewcommand{\arraystretch}{1.10}
\setlength{\tabcolsep}{5.2pt}
\begin{tabular*}{0.94\columnwidth}{@{\extracolsep{\fill}}lccccc@{}}
\toprule
\multirow{2}{*}{Backbone} & \multicolumn{2}{c}{ADD} & \multicolumn{2}{c}{ADD-S} & \multirow{2}{*}{FPS} \\
& AUC & AR & AUC & AR & \\
\midrule
\textbf{CUTIE} & \textbf{31.7} & \textbf{55.1} & \textbf{49.2} & \textbf{73.5} & 55.2 \\
SAM2 & 21.8 & 43.4 & 39.0 & 71.1 & 14.7 \\
OSTracker & 19.3 & 36.9 & 31.9 & 54.3 & \textbf{81.3} \\
\bottomrule
\end{tabular*}
\vspace{-8pt}
\end{table}

OSTracker improves processing speed, reaching 81.3 FPS, but its ADD/ADD-S accuracy degrades noticeably. SAM2 preserves relatively high ADD-S AR, but it is substantially slower. This supports the design of using a memory-based VOS backbone as the 2D observation stream in RRTrack, since it provides more reliable target support for downstream 6D refinement while remaining efficient for online tracking.

\subsubsection{Effects of Key Components}

This study removes recovery-related components to understand their importance after target disappearance. It is evaluated on 72 annotated lost--reappear events spanning pick-and-place manipulation, agile flight, and Dingo navigation, with partial occlusion, full occlusion, and out-of-view disappearance. Under otherwise identical conditions, we test variants by removing the recovery module, pose-gated memory control, and adaptive thresholding, respectively. In the variant without the memory strategy, CUTIE falls back to its automatic memory update with a 5-frame update interval. Table~\ref{tab:ablation_success_overall} reports the ablation. A recovery is considered successful when the tracker re-establishes a pose whose ADD-S point-level AR at $0.1d$ is at least 0.9 after reappearance. The ADD and ADD-S values follow the event-level post-recovery AR defined in the metric section: they average point-level AR over the 10-frame window after the first successful recovery and assign zero to unrecovered events. Therefore, the reported values jointly reflect recovery frequency and the geometric quality of the recovered pose. Delay denotes the median successful recovery delay in frames. FPS uses the equal-subset average over the three synthetic subsets, and higher is better for all metrics.

\begin{table}[!htbp]
\vspace{-8pt}
\centering
\small
\caption{Module-removal and strategy ablation.}
\label{tab:ablation_success_overall}
\renewcommand{\arraystretch}{1.10}
\setlength{\tabcolsep}{4.8pt}
\begin{tabular*}{\columnwidth}{@{\extracolsep{\fill}}lccccc@{}}
\toprule
Ablation setting & Success & ADD & ADD-S & Delay & FPS \\
\midrule
\textbf{Full model} & \textbf{0.722} & \textbf{0.398} & \textbf{0.671} & 2.0 & 55.2 \\
w/o dual-bank matcher & 0.681 & 0.326 & 0.629 & \textbf{1.0} & 63.2 \\
w/o memory strategy & 0.681 & 0.341 & 0.622 & 2.5 & 61.3 \\
w/o adaptive threshold & 0.639 & 0.379 & 0.573 & 9.0 & 62.1 \\
\bottomrule
\end{tabular*}
\vspace{-8pt}
\end{table}

Table~\ref{tab:ablation_success_overall} shows that the full model achieves the best recovery-quality trade-off, with success rate 0.722, failure-aware ADD 0.398, ADD-S 0.671, median successful recovery delay 2.0 frames. Removing individual components gives a small FPS increase, although all variants remain efficient and operate around 60 FPS. Without the recovery module, success rate drops to 0.681 and failure-aware ADD/ADD-S drops to 0.326/0.629, showing that the recovery branch improves both the frequency of valid re-establishment and the geometric quality after disappearance. Without pose-gated memory control, success rate also decreases to 0.681 and failure-aware ADD/ADD-S drops to 0.341/0.622. The fixed-threshold variant remains competitive in ADD, with success rate 0.639 and ADD/ADD-S 0.379/0.573. Its median successful recovery delay increases sharply to 9.0 frames. This indicates that a fixed threshold can still accept some geometrically accurate recoveries, but it tends to require more frames after reappearance, likely because a single confidence cutoff cannot adapt to the large appearance and viewpoint changes across different disappearance events. The adaptive threshold improves recovery responsiveness and temporal consistency while preserving the final pose quality of successful recoveries. These results indicate that recovery, memory control, and adaptive thresholding each contribute to stable recovery after disappearance.

\section{Conclusion}

This paper presented RRTrack, an efficient and robust tracker for object 6D pose tracking in dynamic and occluded scenarios. By closing the loop between memory-based 2D tracking and 6D pose refinement, RRTrack used rendered-mask agreement to verify propagated masks, regulate memory updates, and trigger correction when drift occurs. To handle disappearance--reappearance events without external re-initialization, RRTrack further introduced a DINOv2-based dual-bank recovery module for stable state transitions. Experiments on the proposed synthetic benchmark demonstrated that RRTrack improved equal-subset mean ADD-S AR over recent 6D pose tracking baselines while achieving 55.2 FPS. In future work, we will explore symmetry-aware and probability-distribution-based pose estimation to further improve recovery accuracy for symmetric objects.

\bibliographystyle{IEEEtranTIE}
\bibliography{refs}

\end{document}